\documentclass{article}

\PassOptionsToPackage{numbers, compress}{natbib}

\usepackage[preprint]{neurips_2025}

\usepackage[utf8]{inputenc} 
\usepackage[T1]{fontenc}    
\usepackage{url}            
\usepackage{booktabs}       
\usepackage{amsfonts}       
\usepackage{nicefrac}       
\usepackage{microtype}      
\usepackage{xcolor}         
\usepackage{listings}
\usepackage{csquotes}
\usepackage{graphicx}
\usepackage{subcaption}

\title{Curiosity by Design: An LLM-based Coding Assistant Asking Clarification Questions}

\author{Harsh Darji \\
University of Alberta\\
Camrose, Canada \\
\texttt{hdarji@ualberta.ca} \\
\And
Thibaud Lutellier \\
University of Alberta\\
Camrose, Canada \\
\texttt{lutellie@ualberta.ca} \\
}

\begin{document}

\maketitle
\definecolor{emerald}{rgb}{0.31, 0.78, 0.47}

\newcommand{\todoc}[2]{{\textcolor{#1}{\textbf{#2}}}}
\newcommand{\todoblack}[1]{{\todoc{black}{\textbf{[[#1]]}}}}
\newcommand{\todored}[1]{{\todoc{red}{\textbf{[[#1]]}}}}
\newcommand{\todogreen}[1]{\todoc{emerald}{\textbf{[[#1]]}}}
\newcommand{\todoblue}[1]{\todoc{blue}{\textbf{[[#1]]}}}
\newcommand{\todoorange}[1]{\todoc{orange}{\textbf{[[#1]]}}}
\newcommand{\todobrown}[1]{\todoc{brown}{\textbf{[[#1]]}}}
\newcommand{\todogray}[1]{\todoc{gray}{\textbf{[[#1]]}}}
\newcommand{\todopurple}[1]{\todoc{purple}{\textbf{[[#1]]}}}
\newcommand{\todopink}[1]{\todoc{magenta}{\textbf{[[#1]]}}}
\newcommand{\todocyan}[1]{\todoc{cyan}{\textbf{[[#1]]}}}
\newcommand{\todoviolet}[1]{\todoc{violet}{\textbf{[[#1]]}}}
\newcommand{\todo}[1]{\todoorange{TODO: #1}}

\newcommand{\thibaud}[1]{\todoblue{Thibaud: #1}}

\newcommand{\harsh}[1]{\todobrown{Harsh: #1}}

\newcommand{\code}[1]{\texttt{\small #1}} 
\renewcommand{\todoc}[2]{\relax}
\begin{abstract}
Large Language Models (LLMs) are increasingly used as coding assistants. However, the ambiguity of the developer's prompt often leads to incorrect code generation, as current models struggle to infer user intent without extensive prompt engineering or external context. This work aims to build an LLM-based coding assistant that mimics the human code review process by asking clarification questions when faced with ambiguous or under-specified queries.

Our end-to-end system includes (1) a query classifier trained to detect unclear programming-related queries and (2) a fine-tuned LLM that generates clarification questions. 
Our evaluation shows that the fine-tuned LLM outperforms standard zero-shot prompting in generating useful clarification questions. Furthermore, our user study indicates that users find the clarification questions generated by our model to outperform the baseline, demonstrating that our coding assistant produces more accurate and helpful code responses compared to baseline coding assistants. 

\end{abstract}
\section{Introduction}

Large Language Models (LLMs) have demonstrated remarkable potential as coding assistants by translating informal natural language (NL) intents into functional code (code generation)~\cite{sergeyuk2025using}. However, the inherent ambiguity of NL makes it challenging to ensure that the generated code fully aligns with the user's intended functionality.

Much work highlights the current limitations of LLMs, especially regarding their intent understanding capacities, which may lead to vague, incorrect, or hallucinated answers~\cite{mu2023clarifygpt, zhang2024llm}. Specifically, prior efforts to improve intent specification in coding assistants include prompt engineering techniques and the inclusion of external structured information, such as test cases~\cite{zhang2024plum}. While these strategies enhance the clarity of the prompt and the quality of the AI-generated answer, they place the burden on the user, who needs to guess how much input, test cases, or documentation to supply—an approach that can be both costly (due to prompt length constraints) and impractical. Moreover, such strategies do not fully resolve ambiguity and may introduce irrelevant or distracting information.

A more natural approach mirrors how human assistants and pair programming work: by asking follow-up questions when presented with unclear input. Prior research suggests that dialog-style refinement—where an LLM asks clarification questions—can improve outcomes in code generation tasks~\cite{mu2023clarifygpt, barke2023grounded, ross2023programmer}. This mirrors real-world interactions on platforms like StackOverflow, where clarification often precedes useful answers. However, existing systems rely on zero-shot or few-shot prompting strategies to simulate such behavior. These methods lack robustness since they are not explicitly trained to detect and handle ambiguous input.

\subsection{Motivating Example}

\begin{figure*}[t]
\centering
\includegraphics[width=0.40\textwidth]{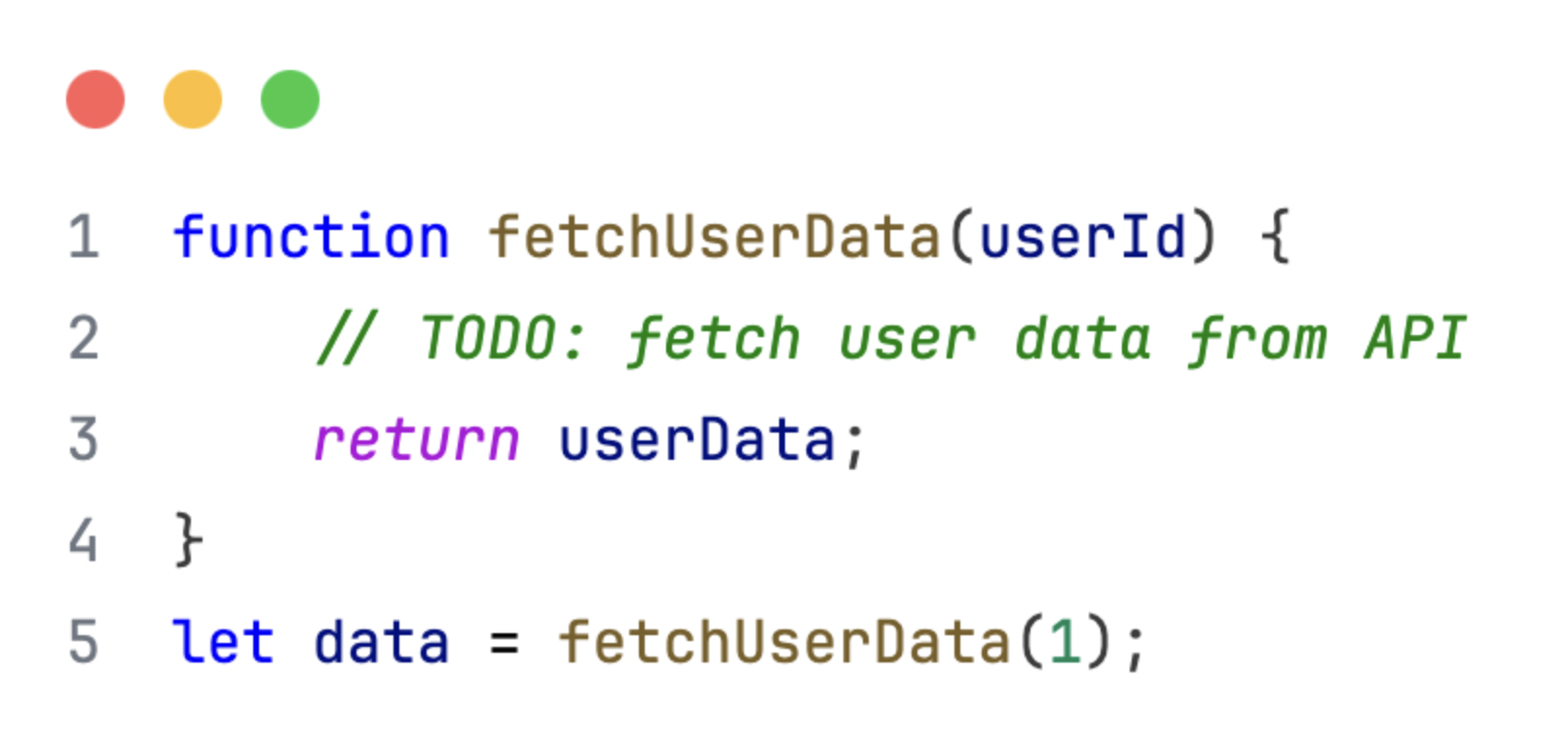}
\caption{Example of under-specified prompt to an LLM.}
\label{fig:motivating}
\end{figure*}

Figure~\ref{fig:motivating} is a typical example of a prompt that a JavaScript software developer might submit to an LLM for code generation. While the prompt appears reasonable, it is under-specified. The developer implicitly expects the LLM to address the TODO comment by generating the code required to fetch user data. 
However, when this prompt is sent to our baseline model, Google's Gemma-3-1B-IT~\cite{team2025gemma}, the model fails to produce the expected code. Instead, it explains in English the purpose of this function. This highlights that the LLM did not correctly capture the user's intent. Instead of generating an unhelpful description of the code, it would have been more effective if the LLM recognized the ambiguity of the prompt and asked the user for additional information. 

\subsection{Our Work}

We propose a new end-to-end coding assistant that asks clarification questions when input is ambiguous or under-specified, mirroring how an experienced human developer might seek to better understand the intent behind a vague coding request. Our solution has two key components; (1) an intent clarity classifier that detects under-specified coding prompts; and (2)a fine-tuned LLM that generates clarification questions before producing a final code answer.

These two components are combined into a single pipeline. When a user submits a natural language coding prompt, the system first applies the intent clarity classifier to determine whether the prompt is sufficiently clear for immediate code generation. If the classifier predicts the prompt is under-specified, the system defers code generation and instead uses the fine-tuned LLM to generate a clarification question. Only after the user provides a follow-up clarification does the assistant proceed to generate the final code output. If the original input is deemed clear, the assistant skips the clarification step and generates code directly. Our evaluation aims to answer two research questions:

\noindent\textbf{RQ1:} Can our new fine-tuned clarification module generate better clarification questions than a standard LLM baseline?

\noindent\textbf{RQ2:} Does our approach yield more precise and useful final code responses than standard LLM pipelines?

Our paper makes the following contributions:

\textbf{Contribution 1.} We fine-tune two new AI models with two new datasets: (1) a coding intent clarity classifier and (2) a coding clarification question generator tailored to under-specified coding queries.

\textbf{Contribution 2.}  We integrate these models into a coding assistant that prompts the user for clarification before answering coding questions.

\textbf{Contribution 3.}  Our evaluation shows that users prefer the final answer provided by our model over the baseline in the vast majority of cases (82\% for precision and focus, 78\% for contextual fit, and 80\% for answer faithfulness).

Our code, datasets, user study templates, and models are publicly available in supplementary material.

\section{Related Work}
\label{related}

Our work is at the intersection of natural language processing and software engineering. We group related work into three different categories: General Conversational Assistants, Coding Assistants, and Intent Clarification.

\subsection{General Conversational Assistant}

Conversational assistants leveraged LLMs to deliver more engaging, adaptive, and context-aware interactions in multiple domains~\cite{chow2024generative,park2023thinking,dong2023towards,mororo2025providing}. Gu et al.\cite{gu2023effectiveness} explored how prompt-based control over personality dimensions can yield distinct conversational styles. Similarly, Zhang et al.\cite{zhang2023survey} surveyed the evolution of conversational search systems, highlighting how LLMs enable complex, multi-turn interactions through modules like query reformulation, clarification, and response generation—laying the groundwork for naturalistic and informative dialogue systems. While such assistants has been well research and are becoming common use, the challenge of unclear or underspecified queries and user intent has risen as a limitation of such assistants.

\subsection{Coding Assistants}

Much work explores how LLMs can be integrated into programming environments to support tasks such as code generation, understanding, debugging, and planning~\cite{wong2023natural,khojah2024beyond,zheng2025towards,liang2024large,kabir2024stack,belzner2023large,paradis2024much}. 
PairCoder~\cite{zhang2024pair,ross2023programmer} draws from pair programming to coordinate two LLM agents. Similarly, ROBIN~\cite{bajpai2024let} introduces a specialized debugging assistant that uses an investigate-and-respond interaction model within GitHub Copilot Chat. 

Others aim to ground LLM assistants in the user’s workflow~\cite{sergeyuk2024ide,fakhoury2024llm,tao2024codelutra,mathews2024test,jin2024can,sergeyuk2025using}. TiCoder~\cite{fakhoury2024llm} introduces a test-driven, interactive loop that supports intent clarification via tests. In a similar spirit, Mathews et al.~\cite{mathews2024test} show that providing test cases upfront can improve the accuracy of LLM outputs. Meanwhile, CodeLutra~\cite{tao2024codelutra} emphasizes learning from both correct and incorrect attempts using preference-guided refinement, improving smaller open-source models' performance without extensive retraining. Our work differs from prior research by focusing on training LLMs to ask clarification questions, rather than assuming user intent can only be inferred with test cases.

Several surveys and empirical studies~\cite{wong2023natural, liang2024large, belzner2023large, nguyen2024beginning,kazemitabaar2023novices,gu2024data,khojah2024beyond, kabir2024stack, jin2024can} emphasize both the promise and limitations of LLMs in code generation, particularly in educational and professional settings. Many of these studies report struggles with intent misalignment and overly literal interpretations of prompts, supporting our motivation for enabling clarification capabilities.

Several studies analyze how programmers engage with AI coding assistants~\cite{barke2023grounded,etsenake2024understanding,chen2024need,gu2024data}. These articles highlight the variability in human-LLM interaction patterns and the need for systems to adapt to individual users and contexts. Our work is built on these conclusions that coding assistants need additional custom information depending on the intent of the user by forcing the assistant to ask clarification questions. 

\subsection{Intent Clarification}

How to handle underspecified queries has been recognized as a challenge in the broader NLP literature~\cite{braslavski2017you, rahmani2023survey, keyvan2022approach, zamani2023conversational}. Foundational datasets such as CLARQ~\cite{kumar-black-2020-clarq} and its recent extension CLARQ-Gen~\cite{gan2024clarq} pair user questions with targeted follow-up queries, and sequence-to-sequence models have been trained to generate such clarification turns. Beyond generic QA, Guan et al.\cite{guan2024large} show that domain knowledge-aware prompts can help LLMs “connect the dots” when generating test cases for deep learning optimizer bugs. Smith et al.\cite{smith2024prompting} leverage “explain in plain English” interactions to turn student descriptions into executable code, using LLM-generated feedback loops to surface and resolve misunderstandings. Yang et al.~\cite{yang2024prompt} apply prompt-based workflows to transform unstructured building code requirements into formal logic clauses, effectively clarifying regulatory intent with minimal training data.

Recent work integrates clarification into code-generation pipelines. ClarifyGPT\cite{mu2023clarifygpt} finds that simply prompting models like GPT-3 to ask a question when uncertain can boost coding task success, but operates in a zero-shot fashion without any model retraining. Kouemo et al.\cite{kouemo2024chain} introduce a chain of targeted verification questions focused on AST nodes to catch and repair bugs in Copilot-generated code, using clarification as a self-refinement mechanism. Endres et al.~\cite{endres2024can} explore whether LLMs can translate informal natural-language intents into formal method postconditions. While these works demonstrate the power of prompting for interactive error detection and specification recovery, they still rely on handcrafted or zero-shot prompts rather than models trained end-to-end.

\section{Approach}
\label{method}

\begin{figure*}[t]
\centering
\includegraphics[width=0.90\textwidth]{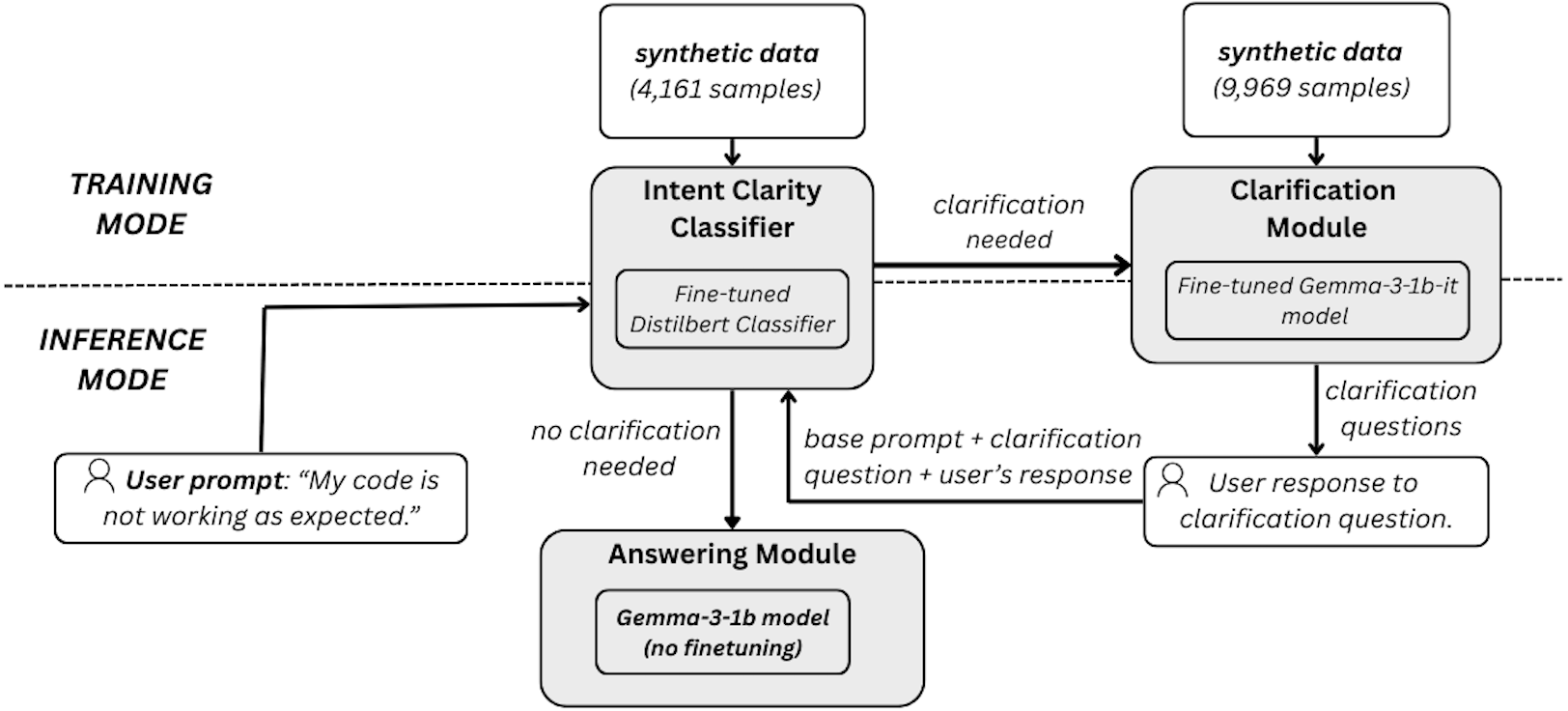}
\caption{Overview of our approach}
\label{fig:overview}
\end{figure*}

This section presents our proposed approach. Section~\ref{over} presents an overview of our system, Section~\ref{intentClassifier} describes our Intent Clarity Classifier, Section~\ref{clarif} presents the Clarification Module, and Section~\ref{answer} describes the Answering Module we use.

\subsection{Overview}
\label{over}

Figure~\ref{fig:overview} provides an overview of our approach. Our system consists of two main phases: a training phase and an inference phase. During training, we fine-tune two components: (1) a DistilBERT-based classifier~\cite{sanh2019distilbert} trained to detect whether a code-related prompt is under-specified (i.e., the prompt's intent is too vague), and (2) a Gemma-3-1B-IT model~\cite{team2025gemma} fine-tuned to generate clarification questions for under-specified prompts. These components are then integrated into an end-to-end pipeline that mimics the behavior of a conversational coding assistant.

At inference time, when a user submits a prompt (e.g., a developer asking a code-related question), the classifier first determines whether the intent behind the prompt is clear (i.e., whether there is sufficient specification to proceed directly with answer generation). If the intent is clear, the system routes the prompt to the code-generation module. However, if the prompt is under-specified, the system calls the clarification-question generator, which generates relevant questions to elicit missing specifications.
The user's response, along with the original prompt and the generated clarification question, is then re-evaluated by the classifier to determine whether the specifications are now clear. If so, the system proceeds to generate the final code response; if not, the clarification loop continues until the prompt is sufficiently clear or a threshold is reached.  

\begin{figure*}[htbp]
\centering
\includegraphics[width=0.90\textwidth]{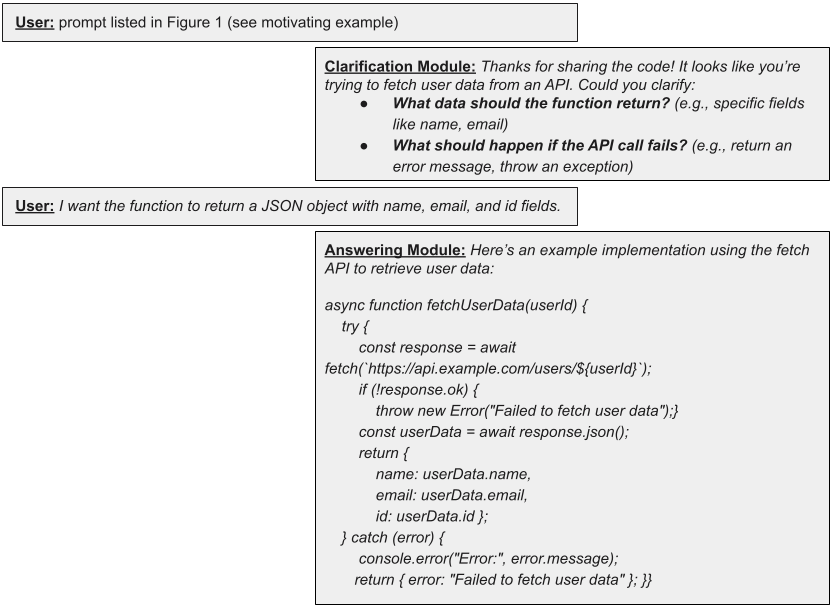}
\caption{Clarification Questions and final answer for the prompt in Figure~\ref{fig:motivating}.}
\label{fig:clarif}
\end{figure*}

For example, the ambiguous prompt in 
Figure~\ref{fig:motivating} is given as input to the Intent Clarity Classifier which identifies it as under-specified. The prompt is then forwarded to the Clarification Module which produces the two clarification questions shown in Figure~\ref{fig:clarif}. The user can respond to these questions (in this case, choosing to answer only the first). The complete dialog, including the original prompt, generated questions, and user responses, is sent back to the Intent Clarity Classifier. This time, the classifier determines that the updated prompt is sufficiently specified and passes it to the Answering Module which, successfully generates the expected code.

We describe below the three main components of our end-to-end system and how we built them.

\subsection{Intent Clarity Classifier}
\label{intentClassifier}

The \emph{Intent Clarity Classifier} (hereafter referred to as \emph{Intent Classifier}) automatically determines whether an incoming user prompt is sufficiently specified for direct response generation or requires follow‐up questions to resolve ambiguity. It consists of a fine-tuned DistilBERT model. We chose DistilBERT due to its smaller size compared to other transformer-based classifiers, ensuring that our pipeline's overhead is minimal. We also employ a 4-point scale for classifying the clarity of user prompts instead of a binary approach, in alignment with previous work~\cite{aliannejadi2020convai3}.

The baseline DistilBERT model is fine-tuned on 
synthetic data generated with GPT-4o-mini, consisting of 4,161 prompt–clarity examples. Out of the 5,000 training samples we attempted to generate, only 4,161 were parseable due to time out in the API calls or LLM hallucinations.
We chose to use synthetic data for several key reasons. First, generating data synthetically using prompt engineering allowed us to create a dataset that was both clean and precisely tailored to the requirements of our study. Second, while we could have data-mined real-world data from StackOverflow or GitHub public issue trackers, this real-world data is in practice extremely noisy and would require a tremendous amount of work to annotate. Even if such data were available, previous work showed that human annotations are prone to error and can have a significant impact on the quality of the data~\cite{zhao2024challenges}. In addition, previous work~\cite{andukuri2024star,malaviya2024contextualized} showed that using synthetic data to fine-tune LLMs can be successful and appropriate in contexts where real data is hard to obtain or noisy. We provide the exact prompt used to generate data in the Appendix.

During inference, prompts detected as well-specified are sent directly to the \emph{Answering module}, while under-specified prompts are routed to the \textit{Clarification Module} to generate follow‑up questions.

\subsection{Clarification Module}
\label{clarif}

The Clarification Module constitutes the second stage of our intent refinement pipeline. It is called when the Intent Classifier identifies a user query as under-specified. 

We selected Gemma-3-1B-IT~\cite{team2025gemma} as our base model for this module to ensure that the question generator would be both scalable and relevant to real-world LLM deployments. Gemma-3-1B-IT represents a moderately sized LLM, making it a practical testbed for parameter-efficient fine-tuning and a realistic proxy for production-scale systems. This choice allowed us to evaluate the effectiveness of our approach on a model size commonly encountered in industry, balancing computational feasibility with representativeness.

Similar to the Intent Classifier training, we fine-tuned the Gemma-3-1B-IT model on a synthetic dataset of 9,969 prompt–clarification pairs generated using OpenAI's GPT-4o-mini. Out of the 10,000 training samples we attempted to generate, 31 failed to generate due to timeouts in the API call.

The dataset comprises two principal categories: (1) code-only prompts that are under-specified with respect to user intent, where it is unclear whether the user seeks a code review, bug fix, optimization, new feature, or something else entirely; and (2) natural language prompts that are code-related but do not include code snippets, such as “I need help connecting to a database. My code keeps throwing errors.” This design ensures coverage of both ambiguous code snippets and vague, code-related queries, reflecting the diversity of real-world developer requests. The prompts used to generate the data are provided in the Appendix.

\subsection{Answering Module}
\label{answer}

The Answering Module consists of a standard coding assistant. In our evaluation, we chose to use the default Gemma-3-1B-IT model for scalability of our experiments and to better control the evaluation since our Clarification Module is a fine-tuned version of Gemma-3-1B-IT. In practice, this could be generalized to any coding assistant.
\section{Evaluation}
\label{evaluation}

Our evaluation aims at answering the following research questions:

\noindent\textbf{RQ1:} Can our new fine-tuned Clarification Module generate better clarification questions than a standard LLM baseline?

\noindent\textbf{RQ2:} Does our approach yield more precise and useful final code responses than standard LLM pipelines?

\subsection{Experimental Settings}
\label{setup}

\smallskip\noindent\textbf{Fine-Tuning Computational Settings:} 
For fine-tuning the Gemma-3-1B-IT model, we used a high-performance computing cluster. Jobs were managed with the following configuration: 4 NVIDIA Tesla P100 GPUs per job, 64 GB RAM, 8 CPU cores, and a maximum walltime of 48 hours. 

\smallskip\noindent\textbf{Inference Computational Settings:}
For training the Intent Classifier and testing the fine-tuned Gemma-3-1B-IT model, we used Google Colaboratory’s free-tier environment, which provides access to NVIDIA Tesla T4 GPUs. Each session was provisioned with a single Tesla T4 GPU (16 GB GDDR6 memory, 2560 CUDA cores, 320 tensor cores), 16 GB system RAM. 

\smallskip\noindent\textbf{Gemma-3-1B-IT Training Settings:}
We employed mixed-precision training and parameter-efficient fine-tuning (LoRA) to optimize resource utilization. Robust error handling and throttling logic were implemented during data generation and training to handle API rate limiting and transient failures, which accounts for minor discrepancies in dataset sizes and training durations.

\subsection{RQ1 Setup}
\label{rq1}

\noindent\textbf{Objective}: RQ1 aims to answer whether our new model can generate better clarification questions than a one-shot prompting baseline. 

\smallskip\noindent\textbf{Study Design:} We answer this question with a user study as the ``quality" of a generated answer is subjective and challenging to measure with standard metrics such as precision and recall. The study was conducted using a within-subjects, pairwise A/B comparison design. Participants were computing science undergraduate research assistants with 2 to 4 years of coding experience. Each participant reviewed 10 interactions, for a total of 100 unique interactions evaluated.

We synthesized the evaluation dataset using GPT-4o-mini with the same prompt template as the training dataset to get an initial set of vague queries. We then manually ensured that each prompt lacked sufficient detail for direct resolution and was representative of real-world queries that typically require clarification in conversational coding assistants. For each prompt, we generated clarification questions using our Clarification Module and the baseline model. 

The clarification questions from both models are then paired for each prompt. To ensure unbiased evaluation, the order of presentation (Clarification Questions A vs. B) was randomized for every prompt and participant. Each participant received a Google Doc containing evaluation instructions and 10 pairs of sample to rate. An instruction sheet is provided in the Appendix. For each prompt, participants were asked to rate each set of clarification questions on three 5-point Likert scales using the following criteria:

\textbf{Precision and Focus}: How exactly the clarification question zeroes in on the missing detail or ambiguity in the original prompt. For example, responses providing examples to
nudge the user to ask a more detailed question to the model.

\textbf{Immediate Editability}: How easily the original prompt could be rewritten using only the clarification question.

\textbf{Contextual Fit:} How well the question aligns with the intent of the original programming task.

\subsection{RQ1 Findings}
\label{rq1findings}

\begin{figure}[htbp]
    \centering
    \begin{subfigure}[b]{0.45\textwidth}
        \includegraphics[width=\textwidth]{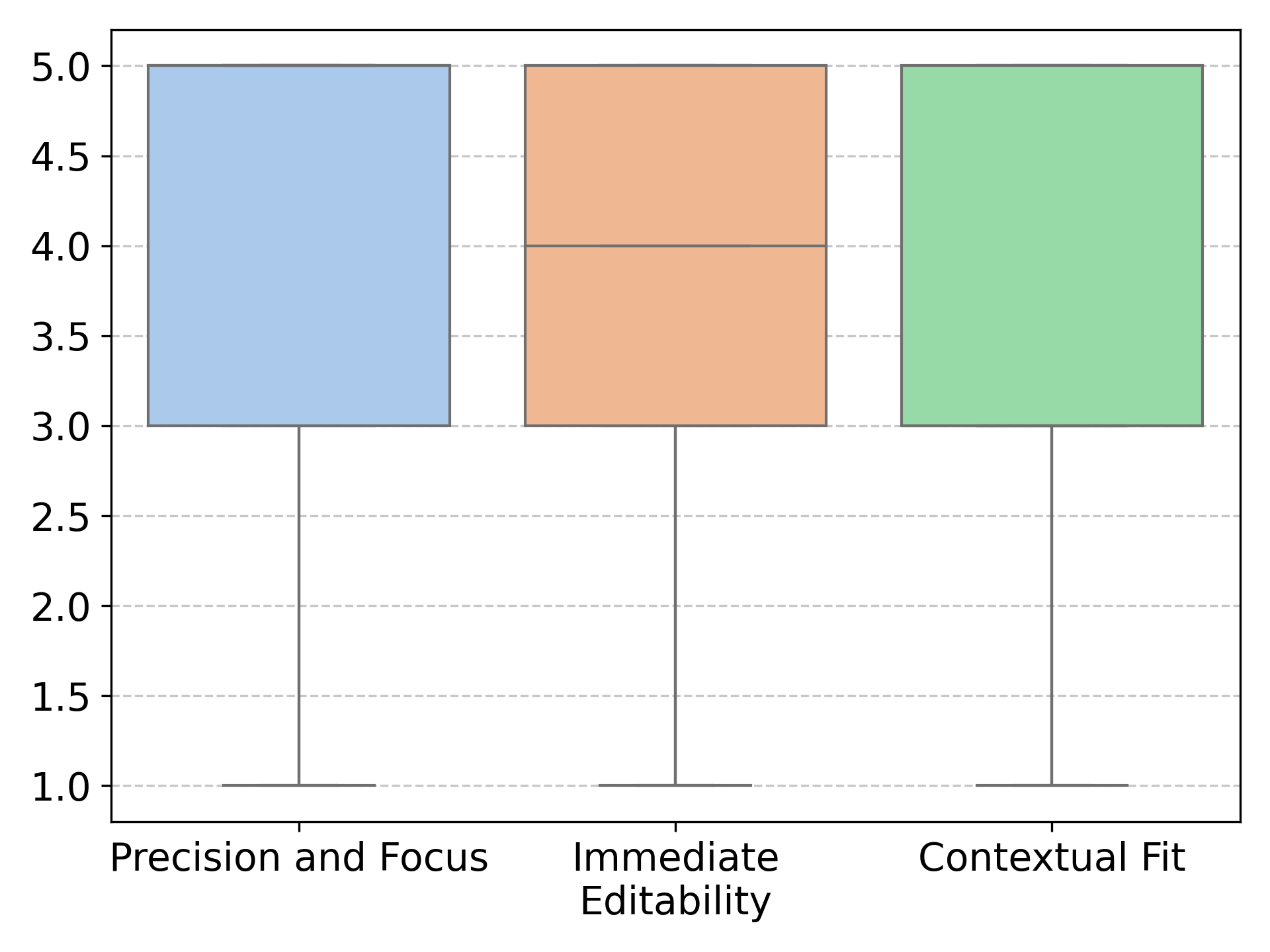}
        \caption{RQ1 Results}
        \label{fig:sub1}
    \end{subfigure}
    \hfill
    \begin{subfigure}[b]{0.45\textwidth}
        \includegraphics[width=\textwidth]{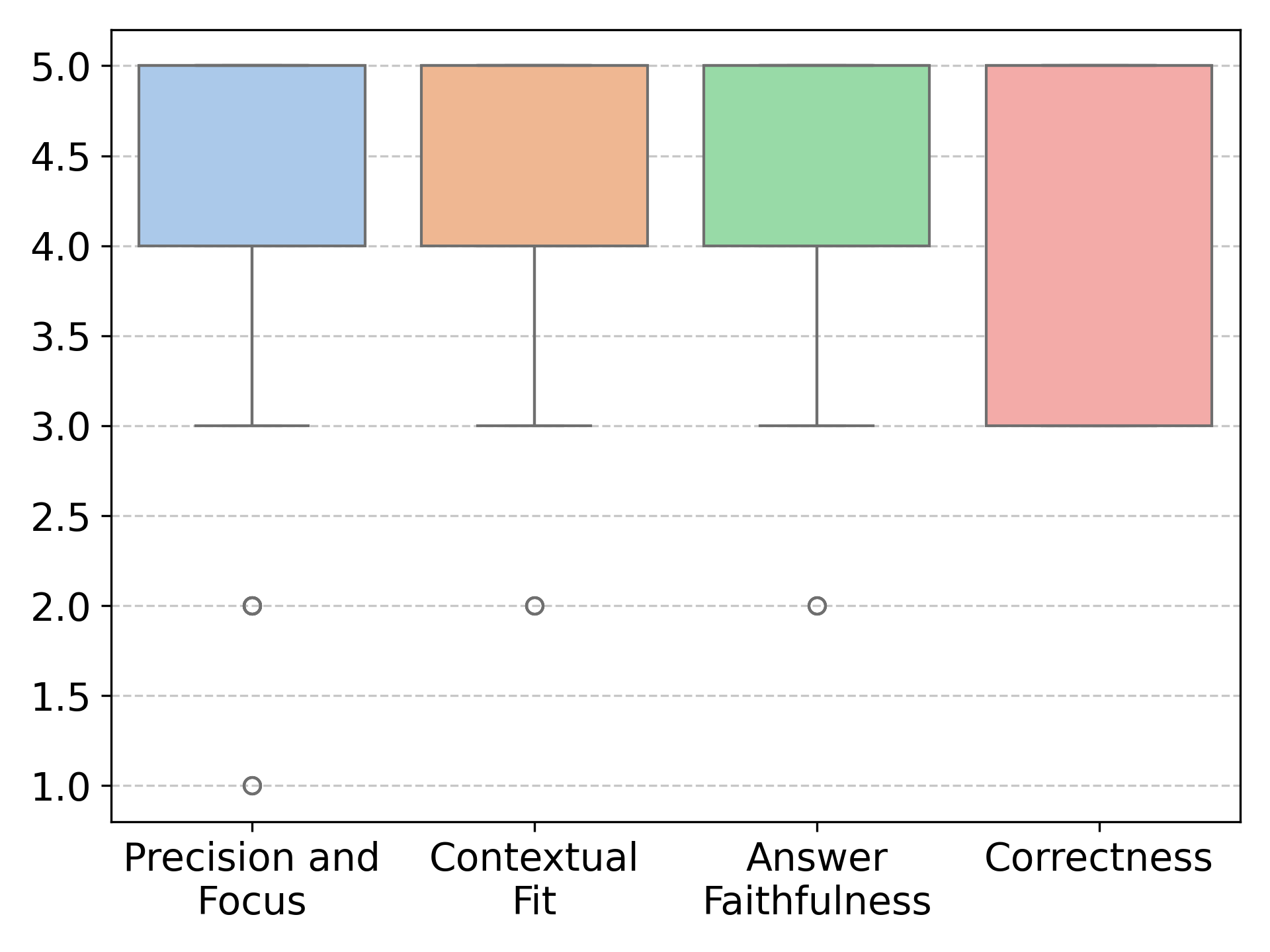}
        \caption{RQ2 Results}
        \label{fig:sub2}
    \end{subfigure}
    \caption{User Study Results}
    \label{fig:main}
\end{figure}

Figure~\ref{fig:sub1} shows box-plot results of the RQ1 user study for the three metrics. A rating of 1 indicates that the user strongly preferred the baseline while a rating of 5 indicates that the user rated the questions generated by our approach as strongly preferable to the baseline. A rating of 3 indicates that the user believes there are no differences between the questions generated by the baseline and our approach. 

We conducted statistical tests for all metrics with a null hypothesis that there are no differences between the two proposed answers (i.e., the user rating should be 3) and all the results rejected the null hypothesis with a p-value $<0.001$. We also measured the effect size using Cohen's d and observed a medium effect size for precision and focus ($d = 0.61$) and a small effect size for other metrics ($0.3 < d < 0.5$).

In terms of precision and focus, users favored our approach (rating of 4 or 5) in 68\% of cases, with a mean rating of 3.9. In terms of immediate editability, users favored our approach only in 53\% of cases, with a mean of 3.55. However, they considered our approach equivalent or better (rating of 3, 4, and 5) in 80\% of cases. Finally, in terms of contextual fit, users considered our approach better in 45\% of cases and equivalent or better in 87\% of cases, with a mean rating of 3.56.

These findings demonstrate a clear user preference for the questions generated by our approach over those from the baseline, particularly in terms of precision and focus. The strong statistical significance ($p< 0.001$) across all metrics and the large effect sizes (Cohen’s d $>0.8$) suggest that these differences are meaningful. While the advantage was most pronounced in precision and focus, where nearly 70\% of users rated our questions favorably, the results for immediate editability and contextual fit also show that our approach is generally perceived as equal or superior to the baseline.

\noindent\fbox{
\centering
\parbox{0.97\linewidth}{
\textbf{Finding 1}: On average, users significantly preferred questions generated by our approach across all metrics, with 68\% favoring it for precision and focus (mean = 3.9 out of 5), 53\% for immediate editability (mean = 3.55 out of 5), and 45\% for contextual fit (mean = 3.56 out of 5), with all differences statistically significant (p < 0.001) and large effect sizes (Cohen’s d > 0.8).
}}

\subsection{RQ2 Setup}
\label{rq2setup}

\textbf{Objective}: RQ2 evaluates whether our approach yields more precise and useful final code responses than a standard LLM pipeline that receives only the initial vague prompt and generates an answer without seeking additional context or clarification.

\textbf{Study Design}: This user study involved 10 undergraduate research assistants with 2 to 4 years of coding experience to review a sample of 5 chat histories for a total of 50 chat histories. We shortened the number of samples to review compared to RQ1 since the total chat histories are longer than the samples used in RQ1 to limit the impact of reading fatigue on our results.

Each sample includes two chat histories: one from our pipeline (with the initial under-specified prompt, clarification question, simulated user response, and final answer) and one from the baseline (with just the prompt and baseline answer). We chose to simulate user responses to the clarification questions to reduce user burden and increase control of the evaluation.

Each participant received a Google Doc with evaluation instructions, rating scales, and pairs of chat histories. To mitigate bias, the order of the baseline and enriched-context responses was randomized per prompt and per participant.
For each prompt, participants were asked to rate each set of responses on 5-point Likert scales using the following criteria:

\textbf{Precision and Focus}: How directly the response targets the question's core issue, using clear guidance, examples, or code.

\textbf{Contextual Fit:} How well the response matches the user's programming goal (e.g., debugging, feature implementation).

\textbf{Answer Faithfulness}: How effectively the response provides actionable steps or code to achieve the user's desired outcome.

\textbf{Correctness}: Accuracy and factual validity of the information, code, or guidance provided, ensuring it is free from errors and aligns with the user's query and intended outcome.

These metrics differ from RQ1 because they target different goals. Immediate editability is only relevant to RQ1 (since there is no prompt refinement required after the final response) whereas answer faithfulness and correctness are critical for RQ2. 
\subsection{RQ2 Findings}
\label{rq2findings}
Figure~\ref{fig:sub2} shows box-plot results of RQ2 user study for the four chosen metrics. A rating of 1 indicates that the user strongly preferred the baseline while a rating of 5 indicates that the users rated the answer generated by our approach as strongly preferable to the baseline. A rating of 3 indicates that the user believes there are no differences between the two generated contents.

We measured statistical tests with the same null hypothesis as in RQ1 and all the results rejected the null hypothesis with a p-value $<0.001$. We also measured the effect size using Cohen's d and observed a very large effect size for all metrics ($>1.2$).

In terms of precision and focus, users favored our approach (rating of 4 or 5) in 82\% of cases, with a mean rating of 4.4. In terms of contextual fit, users favored our approach in 78\% of cases, with a mean of 4.3. For answer faithfulness, users considered our approach better in 80\% of cases, with a mean rating of 4.4. Finally, in terms of correctness, 100\% of cases were deemed at least as correct as the baseline, with our approach producing more correct answers in 66\% of cases (median of 4.3).
 
\noindent\fbox{
\centering
\parbox{0.97\linewidth}{
\textbf{Finding 2}: Users strongly preferred answers generated by our approach over the baseline across all metrics, in 82\% of cases for precision and focus, 78\% for contextual fit, 80\% for answer faithfulness, and 66\% for correctness, all result being statistically significant  (p < 0.001) with very large effect size (Cohen’s d > 1.2).
}}

\subsection{Additional Evaluations}
\label{addEval}

In addition to our two research questions, we performed three additional evaluations to study the performance of our Intent Classifier, the performance of our system using standard NLP metrics (perplexity), and the computational overhead induced by our pipeline.

\noindent\textbf{Intent Classifier Evaluation:} Using cross-validation, the accuracy of the Intent Classifier was 73\% (Precision of 73\% and Recall of 71\%). This is reasonably high for our purpose and aligns with previous ambiguity classifiers' accuracy.  

\noindent\textbf{Quantitative Evaluation:} We quantitatively evaluate our approach by measuring the perplexity of the fine-tuned Gemma-3-1B-IT model on the clarification question generation task. Perplexity is a standard metric in NLP that reflects how well a language model predicts a sequence of tokens, with lower values indicating higher confidence and better fit to the data. Our fine-tuned model achieves a 4\% reduction in perplexity, indicating improved confidence and better adaptation to the clarification question generation task.

\noindent\textbf{Pipeline Overhead:} 
Training the Intent ClarityClassifier required approximately one hour, while fine-tuning the Gemma-3-1B-IT Clarification Module took about four hours on our server. In inference, the main pipeline overhead is introduced by the Clarification Module which takes an average of 133 seconds per prompt to generate clarification questions, the intent classification taking less than 0.1 seconds per prompt.

\section{Discussion}
\label{discussions}

\noindent\textbf{Broader Impact}
There is a risk that the coding assistant provides incorrect advice or generates insecure code that leads to the introduction of vulnerabilities in software~\cite{toth2024llms}. This is a limitation of all coding assistants and users should be aware of such limitations of AI. Previous work~\cite{mohsin2024can} found that adding security patterns in the prompt led to more secure code generation. Anecdotally,  the example in Figure~\ref{fig:clarif} shows that our model asks a security-related question (What should happen if the API call fails?) which may have led to the generation of the try-catch block of code. Further study could assess whether intent clarification such as the one proposed in this work leads to more secure generated code.
 
\subsection{Limitations}

\noindent\textbf{Conclusion Validity:}
Our evaluation includes a user-simulation component, in which a larger LLM responds to clarification questions instead of a user. We chose this setup for our evaluation as it provides scalability and allows us to reduce the burden of the user study on our participants. While it may misrepresent interactions with real users, we mitigate this threat by having participants of our user study rate both the clarifications and the final answer separately. Overall, clarification responses synthetically generated are likely of a lower quality than real user responses (since the LLM may not be aware of the real intent of the initial query), which reduces the quality of our evaluation.

\smallskip\noindent\textbf{Construct Validity:}
Our classifier for detecting under-specified prompts has good, but not perfect accuracy. This does not severely impact the system’s overall utility. In practice, false positives (i.e., generating clarification questions for already clear prompts) are low-cost and do not significantly degrade answer quality. 

\smallskip\noindent\textbf{Internal Validity:}
Our dataset was generated synthetically to ensure high-quality training data. While synthetic prompts offer control and balance, they may not capture the variability of real-world developer queries. We initially collected a dataset of under-specified queries, clarification questions, and answers from the issue threads of the top 1,000 GitHub projects. In practice, this data was too noisy and led to lower model performance. In the future, we will investigate better cleaning and filtering techniques to integrate this data to our synthetic data to extend the generalization of our dataset. 

\section{Conclusion}
\label{conclusion}

We proposed a new end-to-end coding assistant that addresses vague queries by first detecting under-specified prompts and then generating clarification questions before producing a final response. Our results show that the fine-tuned clarification model outperforms standard prompting in producing relevant and helpful clarifications. User studies further confirm a strong preference for our model and pipeline over traditional LLM baselines. These findings suggest that clarification-driven interactions are a promising direction for the future LLM-based coding assistants. By enabling LLMs to ask the right questions, we take a step towards assistants that are not only reactive but collaborative.

\clearpage

\bibliographystyle{ACM-Reference-Format}

\bibliography{main}

\end{document}